\pdfoutput=1

\documentclass[11pt]{article}

\usepackage[preprint]{acl}

\usepackage{times}
\usepackage{latexsym}

\usepackage[T1]{fontenc}

\usepackage[utf8]{inputenc}

\usepackage{microtype}

\usepackage{inconsolata}

\usepackage{graphicx}

\title{Mapping Toxic Comments Across Demographics: \\ A Dataset from German Public Broadcasting}

 \author{Jan Fillies \\ Freie Universität Berlin \\ Fraunhofer FOKUS    \And
         Michael Peter Hoffmann \\ Freie Universität Berlin \And
          Rebecca Reichel \\ MSB Medical School Berlin \AND
        Roman Salzwedel \\ funk - Content-Netzwerk \\ ARD / ZDF\And
        Sven Bodemer\\ funk - Content-Netzwerk \\ ARD / ZDF\And
        Adrian Paschke \\ Freie Universität Berlin \\ InfAI Leipzig  \\ Fraunhofer FOKUS}

\begin{document}
\maketitle
\begin{abstract}
A lack of demographic context in existing toxic speech datasets limits our understanding of how different age groups communicate online. In collaboration with funk, a German public service content network, this research introduces the first large-scale German dataset annotated for toxicity and enriched with platform-provided age estimates. The dataset includes 3,024 human-annotated and 30,024 LLM-annotated anonymized comments from Instagram, TikTok, and YouTube. To ensure relevance, comments were consolidated using predefined toxic keywords, resulting in 16.7\% labeled as problematic. The annotation pipeline combined human expertise with state-of-the-art language models, identifying key categories such as insults, disinformation, and criticism of broadcasting fees. The dataset reveals age-based differences in toxic speech patterns, with younger users favoring expressive language and older users more often engaging in disinformation and devaluation. This resource provides new opportunities for studying linguistic variation across demographics and supports the development of more equitable and age-aware content moderation systems.

\end{abstract}

\section{Introduction}
Social media has become a dominant space for public discourse \cite{gulzar2023role}, yet researchers and policymakers lack access to key demographic data, especially age, that would allow for a nuanced understanding of how different groups communicate online. Age is a particularly important dimension, as it influences language use, topic preferences \cite{eckert2012three}, and susceptibility to or propagation of harmful content. However, most platforms restrict access to such information, and existing toxic speech datasets offer limited support for demographic analysis.

This research addresses that gap by introducing the first large-scale multi-platform, German-language dataset of social media comments annotated for toxicity and enriched with estimated user age ranges. The dataset was developed in partnership with \textit{funk},\footnote{https://play.funk.net/funk} a digital content network operated by German public broadcasters ARD and ZDF. Funk and its subsidiary accounts target users aged 14–29 and, as a content creator on Instagram, TikTok, and YouTube, they have access to anonymized, platform-provided age distribution data, a novel opportunity for analysis.

The dataset comprises 3,024 human-annotated comments, and annotation was extrapolated using large language models (LLMs) to label an additional 30,024 comments. The performed analysis of both datasets reveals significant age-related differences in toxic language use: younger users tend to employ more expressive or sarcastic language, while users aged 31–35 are more likely to produce disinformation and devaluation. Toxicity patterns also vary across platforms, with Instagram showing the highest proportion of toxic content.

This resource enables novel fine-grained demographic analysis of online toxicity and offers a foundation for developing more equitable, age-aware moderation systems.

The main contributions of this research are: 
\begin{enumerate}
\item A novel multi-platform German dataset for toxic speech, comprising 3,024 human-annotated and 30,024 LLM-annotated comments, including age demographics.
\item A comparative evaluation of four LLMs (GPT-3.5, GPT-4o, GPT-4o-mini, and Llama-3-8B) for toxicity annotation.
\item A detailed quantitative analysis of the datasets, highlighting the differences in age groups. 
\end{enumerate}
The dataset is published via Zenodo\footnote{Anonymous Submission} and is available to verified researchers, with active research projects, upon request, and after evaluation by the research partners.

\section{Related Work}
Annotated toxic and hate speech datasets have grown significantly in recent years. While most remain in English (e.g., \cite{Hosseinmardi2015,Gibert2018,ElSherief2018,Wen2022}), efforts have expanded to other languages \cite{sanguinetti-etal-2018-italian,Mandl2019}. Key monolingual datasets include \citet{davidson2017automatedhatespeechdetection}'s 2017 collection of 24,802 English tweets labeled as hate speech, offensive language, or neither. \citet{founta2018largescalecrowdsourcingcharacterization} compiled 80,000 English tweets annotated for abusive behavior in 2018. HateExplain \cite{mathew2021hatexplain} features 20,148 English posts with word-level annotations. More recently, \citet{fillies2023hateful} collected 88,000 English Discord messages with eight hate speech classes.

For this study, German annotated datasets are particularly relevant. An early contribution by \citet{Bretschneider2017} includes 469 German tweets with binary hate labels. Expanding on this, GermEval 2018 \cite{Wiegand2019} features 2,871 tweets classified as offensive. \citet{Mandl2019} introduced a 2019 multilingual dataset with 4,669 German tweets and Facebook comments labeled for hate speech, offensive, and profane language. Similarly, \citet{demus-etal-2022-comprehensive} annotated 10,278 Twitter comment threads with a detailed schema, identifying 10.85\% hateful content. Moving beyond Twitter, \citet{Assenmacher2021} compiled 85,000 news website comments, labeled abuse across seven categories. Recently, \citet{goldzycher2024improvingadversarialdatacollection} released 10,996 texts combining synthetic and news data, with 42.4\% hateful content. Lastly, \cite{keller2025hocon34k} published a 34,223-comment binary hate speech dataset from online discourse related to German newspapers.

As demonstrated, toxic and hate speech datasets employ diverse annotation schemes \cite{Chung2019}, ranging from binary classifications to multi-class hierarchies \cite{Ranasinghe2020}, and even more universal schemes, which are widely used in related tasks like cyberbullying annotation \cite{Sprugnoli2018}. This diversity highlights the flexibility of these schemes in addressing various research challenges.

Recent research on using LLMs for annotating toxic and hate speech datasets shows promise in reducing bias and inter-annotator variability. Studies indicate that GPT-based models can assist in pre-annotation by providing initial labels for human review, cutting time and costs \cite{das2024investigatingannotatorbiaslarge}. Additionally, LLMs' interpretability helps analyze nuanced language structures like sarcasm and implicit hate, which traditional methods often struggle with \cite{roy-etal-2023-probing}.


Although many German datasets are publicly available, they predominantly focus on Twitter, with limited exploration of other platforms such as news outlets and Facebook comments. Platforms like TikTok remain understudied, and datasets containing multiple platforms annotated within a single annotation schema are very rare. No dataset containing age ranges directly provided by the platforms could be identified. This research contributes to the field by introducing the first multi-platform age range annotated dataset for German toxic speech.

\section{Dataset Collection}
\label{data_collection}
The work presents two datasets. The first is a dataset consisting of 3,024 human-annotated comments, and the second is a dataset comprising 30,024 different comments annotated with the support of a LLM, based on the annotations and definitions used in the first dataset. 
Both datasets consist of comments posted under long- and short-format videos shared on funk's main social media channels and their associated accounts on TikTok, Instagram, and YouTube. The dataset includes, but does not specifically highlight, comments that were concealed from the public by funk's content moderation team. 

The platform provided age information for the audience of each creator account. Information on age was not provided by TikTok, but is provided by the other two platforms based on how many followers (Instagram) or how many views (YouTube) fall into one of a set of predefined age groups.\footnote{For YouTube and Instagram these age groups are: 13-17, 18-24, 25-34, 35-44, 45-54, 55-64, 65 and older.} Based on the age group distribution, Funk calculates an average age for each content format (i.e., creator account). It is assumed that the number of followers or views within each age group is evenly distributed across the individual ages within that category. Although the age groups provided by the platform rely on user-reported birth dates, which may be inaccurate, and the groupings per channel are relatively broad, this method remains the most reliable approximation of official age distribution data currently available to researchers, as it offers the only consistent, large-scale, and platform-specific demographic insight accessible in the absence of more precise or independently verified data.



Similar to \citet{waseem-hovy-2016-hateful}, the initial collection of comments was consolidated for relevance on a predefined word list, this list consists of a mix of established word lists and terms collected by the research group out of past projects, see Appendix \ref{wordlist}. The final list contains a range of vocabulary related to toxic speech, it is available on GitHub.\footnote{Annonymous}

Even after this initial filtering, funk receives substantially more than 33,024 comments under its posts over the course of a year (100,000+). Since comments were collected between January 1, 2023, and December 31, 2023, the research selected an equal number of comments per month, distributing them equally among the funk accounts that had available comments during that period. If there were not enough comments from a particular account or month, the research included statements from the same accounts in the adjacent months, ensuring that no statements were selected twice.

After selection, all comments were anonymized and pseudonymized by funk. This process involved a combination of Regular Expressions (Regex) and advanced Named Entity Recognition (NER) techniques to identify emails, IBANs, phone numbers, locations, and private individuals. MD5 hashing with added SALT was employed to pseudonymize all locations and individuals, except for those identified as known politicians from the US, UK, or Germany. The list of politicians was sourced from the EveryPolitician Names project.\footnote{https://github.com/everypolitician/everypolitician-names}

The script used for this process is available on GitHub.\footnote{Annonymous} For transparency, a data statement is presented in Table \ref{datastteament} in Appendix \ref{Data_statement}. It was created following \citet{Gebru2021}.

\section{Annotation Scheme and Guidelines}
\label{Annoattion_scheme}

For the research project, the schema and annotation guidelines were developed in cooperation with content moderators from funk, domain experts on toxic online language, and through an iterative process involving the annotators during annotation, as suggested by \citet{Vidgen2021}. The guidelines are based on the core elements of funk’s content policies,\footnote{https://play.funk.net/netiquette} which reflect their understanding of problematic content. For many general aspects, the taxonomy was inspired by the framework proposed by \citet{fillies2025novel}, which served as a foundational reference for structuring categories.

The annotation scheme consists of 18 labels and two main classes: the target of the toxic language and the type of language. Table \ref{combined-categories} displays the possible labels for each class. The labels “Criticism of Public Broadcasting Fees,” “Suicide,” and “Disinformation” are not necessarily considered toxic or problematic in all cases. “Criticism of Public Broadcasting Fees” was included because understanding criticism of these fees is important to the media outlet. The labels “Suicide” and “Disinformation” were included as they still raise concerns, and funk aims to address these issues within its content moderation efforts.

The guideline provides descriptions for each label along with example reference statements to further illustrate each element. It also emphasizes the importance of context sensitivity, instructing annotators to evaluate comments as they would appear under funk videos, even without precise contextual details. Annotators were informed that multiple labels could be assigned and were instructed to provide a severity rating (e.g., Urgent or Non-Urgent) to reflect the potential harm or the need for immediate moderation. In cases where disinformation was identified, annotators were directed to label the types of toxic content, if any, or to solely mark the disinformation if no hate was detected. The full annotation guidelines can be found on GitHub.\footnote{Annonymous}

                      
\begin{table}
    \centering
    \begin{tabular}{ll}
        \hline
        \textbf{Class} & \textbf{Categories} \\
        \hline
        Target         & Religion, Ethnic/Racial/Nationality, \\
                      & Physical Condition, Gender and Sexual\\ & Identity, Occupation-Based, Critic \\ & of Public Broadcasting Fees  \\ & Class, Other \\
        \hline
        Type           & Violence, Insults, Devaluation, \\
                      & Discrimination, Threats, \\
                      & Disinformation, Suicide, \\
                      & Spam and Scam, Other, \\

        \hline
    \end{tabular}
    \caption{\label{combined-categories}
        Combined list of categories grouped into two classes: "Target" and "Type."
    }
\end{table}

\begin{table}
  \centering
  \begin{tabular}{ccccc}
    \hline
    \textbf{annotator} & \textbf{avg \%}  & \textbf{avg cohen k}  &\textbf{fleiss k} \\
    \hline
    Human & 0.89       & 0.63      &          0.64             \\
    GPT-3.5 & 0.97 & 0.61 & -  \\
    GPT-4o & 0.98 & 0.81 & - \\
    GPT-4o-mini & 0.96 & 0.74 &  - \\
    Llama-3 8B & 0.88 & 0.29 & - \\
    \hline
  \end{tabular}
  \caption{\label{human_inter_anno_agreement}
    Inter-annotator agreements. Human agreement measures consistency between annotators, while model agreement shows how closely a model matches the human consensus.
  }
\end{table}

\section{Annotation}
\subsection{Human Annotation}
The three annotators were tasked with independently annotating each statement. Annotators were also responsible for conducting additional research if they encountered unfamiliar concepts. In such cases, they could leave separate comments for their peers, providing further information or requesting assistance in interpreting certain aspects of the comment. Annotators were asked to flag comments that contained potential personal information. For all problematic statements that were difficult to classify, the annotators met biweekly, discussing problematic cases individually to ensure high-quality annotations and high agreement, as recommended by \citet{Vidgen2021}.


Table \ref{human_inter_anno_agreement} displays the agreement levels obtained between the three annotators in terms of the average percent agreement (avg \%), the average Cohen’s kappa coefficient \cite{cohen1960coefficient} (avg Cohen’s k), and Fleiss’ kappa coefficient \cite{fleiss1971measuring} (Fleiss’ k). The metrics used and the achieved results are in line with similar studies \cite{de2018hate}. A closer break down of the inter-annotator agreement can be Found in Appendix \ref{interanno_break}.

\begin{table}
  \centering
  \begin{tabular}{ccc}
    \hline
    \textbf{Model} & \textbf{Time (min)}           & \textbf{Cost (\$)} \\
    \hline
    GPT-3.5 & 29 (min) 23 (sec) & 0.79   \\
    GPT-4o & 58 (min) 25 (sec) & 4.31  \\
    GPT-4o-mini & 40 (min) 16 (sec)  & 0.36\\
    Llama-3 8B & 36 (min) 68 (sec) &  -  \\
    \hline
  \end{tabular}
  \caption{\label{LLM_cost}
    Annotation time and cost of the 3,024 comment dataset.
  }
\end{table}

\subsection{LLM Annotation}
Four prominent models, GPT-3.5, GPT-4o, GPT-4o-mini, and Llama-3-8B, were evaluated for their prompt-based ability to classify text according to the provided annotation schema.

\citet{das2024offensivelang} designed and evaluated different prompts for LLM-based offensive content detection. This research builds upon their best-performing prompt design, extending it from a binary to a multi-class classification task. The newly developed prompt incorporates the annotation guide and requests classification as follows:

“Based on the following annotation schema:
[Annotation Guideline]
Carefully analyze the comment given by the user for ALL possible categories. If the comment is non-toxic, return 'Non-Toxic' with no annotations. If the comment is toxic, categorize and identify ALL relevant targets and speech types, and assign a severity rating. Response Format: [...]”


Table \ref{human_inter_anno_agreement} displays the Cohen's kappa values for each model’s individual predictions compared to the majority-vote annotations of the human annotators. The results show that Llama-3 8B performs the weakest, while GPT-4o has the highest agreement with the annotators. Notably, both GPT-4o and GPT-4o-mini outperform the average human Cohen's kappa by a substantial margin, meaning they agree with the majority human consensus more consistently than the average individual human annotator does. GPT-3.5 produces an acceptable agreement range. It is important to note that the higher agreement of LLMs reflects consistency with the consensus, not necessarily superior judgment or understanding. This may indicate alignment with dominant patterns in the data, but not deeper contextual reasoning or the ability to resolve ambiguous cases.

Table \ref{LLM_cost} presents the time and costs associated with each LLM-based annotation process. GPT-3.5 was the fastest, while GPT-4o was the slowest. Llama-3-8B was free of charge, whereas GPT-4o-mini was the most cost-effective GPT model.

To compare model performance, three metrics are used. First, accuracy, which measures the proportion of correct predictions but can be misleading in imbalanced datasets. Second, the Macro F1 Score is the average F1 score calculated per class, treating all classes equally. This makes it a more suitable metric for evaluating performance on imbalanced classification tasks. Lastly, the Matthews Correlation Coefficient (MCC), a balanced metric. The advantage of MCC is that it accounts for true negative predictions, unlike the F1 score \cite{chicco2020advantages}. Given the highly imbalanced nature of the dataset, this makes MCC particularly suitable in this context. Table \ref{Perfoamnce_models} displays the performance of each model on the annotated dataset. The results show that GPT-4o performs best across all three metrics, followed by GPT-4o-mini. The low Macro F1 score can be attributed to the high imbalance and sparsity of the dataset.

\begin{table}
  \centering
  \begin{tabular}{p{0.13\textwidth}p{0.06\textwidth}p{0.06\textwidth}p{0.06\textwidth}}
    \hline
    \textbf{Model}  & \textbf{Macro F1} & \textbf{Acc.}   & \textbf{MCC}\\
    \hline
    GPT-3.5  & 0.10 & \textbf{0.97} & 0.51 \\
    GPT-4o& \textbf{0.36} & \textbf{0.97} & \textbf{0.75}\\
    GPT-4o-mini & 0.33 & 0.96 & 0.67\\
    llama-3 8B  & 0.15 & 0.87 & 0.22\\
    \hline
    GPT-4o-mini-fine& \textbf{0.38} & \textbf{0.97} & \textbf{0.78}\\
    \hline
  \end{tabular}
  \caption{\label{Perfoamnce_models}
    Performance of the models on the annotated dataset. The results of GPT-4o-mini-fine were calculated on an evaluation test set.
  }
\end{table}

\section{Fine-tuning and Extrapolation}
Based on performance, cost, and time efficiency, GPT-4o-mini was selected for the annotation of the 30,024 comments. To further improve performance, the model was fine-tuned using the following hyperparameter settings: Epochs: 5, Batch Size: 3, Learning Rate: 0.3. These initial choices were guided by prior work on similar model scales and task types, where smaller batch sizes and moderately aggressive learning rates helped accelerate convergence without overfitting. Five epochs were chosen as a balance between sufficient learning and computational cost, based on preliminary learning curve assessments on a small subset.

Fine-tuning was conducted on a stratified training and test set (90\%-10\%), while the results in Table \ref{Perfoamnce_models} were calculated on a separate holdout set. Further hyperparameter optimization was performed using a simple staging approach, in which small-scale experiments were first conducted to rapidly evaluate different parameter configurations (e.g., learning rate, batch size, and number of training epochs), but these did not surpass the initial settings. The results of these additional experiments are listed in Appendix \ref{fine_tune}. Notably, the fine-tuned model outperforms the GPT-4o model. The training time and costs for the fine-tuning and annotation can be seen in Table \ref{fine_tuning_annoattion}.

The use of GPT-4o-mini illustrates the practicality of LLMs for scaling annotation. Given its short annotation time, low cost, and high agreement with human annotators, it offers an efficient way to extend human-labeled standards to large datasets.

\begin{table}
  \centering
  \begin{tabular}{ccc}
    \hline
     & \textbf{Time (min)}           & \textbf{Cost (\$)} \\
    \hline
    Fine-tuning & 36 (min) 31 (sec) & 1.49   \\
    Annotation &  7 (h) 45 (min) 5.68 (sec) & 4.04 \\
    \hline
  \end{tabular}
  \caption{\label{fine_tuning_annoattion}
     Time and Cost for Training of the fine-tuned model and annotation of the 30,024 statements.
  }
\end{table}




\section{Evaluation of Human Annotated Dataset}
The 3,024 statements were posted by 2,951 unique users under 2,112 unique posts from 141 different accounts, averaging 89.17 accounts per month. A total of 1,008 statements were selected across the three platforms: YouTube, Instagram, and TikTok, which is 82 statements per platform per month. 

Table \ref{label_distribution_combined} presents the distribution of labels based on the three annotations combined via majority voting. The results indicate that the majority of entries (83.30\%) are labeled as "non-toxic." Among the types of toxic speech, "Insults" (6.51\%), "Devaluation" (5.5\%), and "Disinformation" (2.78\%) are most represented. In the target group, "Occupation-Based Hate" (1.95\%), "Gender and Sexual Identity" (1.88\%), and Rundfunkgebühren (Broadcasting fees) (1.88\%) are highest ranked. 
Less frequent labels, including threats, violence, and suicide-related content, each account for less than one percent. This distribution highlights the imbalanced nature of the dataset, which is a common characteristic in similar studies \cite{fillies2023hateful}.

\begin{table}
  \centering
  \begin{tabular}{p{0.1\textwidth}p{0.06\textwidth}p{0.06\textwidth}p{0.06\textwidth}p{0.06\textwidth}}
    \hline
    \textbf{Label} & \textbf{Hum. Count} & \textbf{Hum. (\%)} & \textbf{LLM Count} & \textbf{LLM (\%)} \\
    \hline
    non-toxic & 2519 & 83.30 & 23123 & 77.02 \\
    Insults & 197 & 6.52 & 3180 & 10.59 \\
    Deval. & 166 & 5.50 & 2403 & 8.00 \\
    Disin. & 84 & 2.78 & 1250 & 4.16 \\
    Discri. & 71 & 2.35 & 617 & 2.06 \\
    Occup.&59 & 1.95 & 689 & 2.29 \\
    Gen./Sex. & 57 & 1.88 & 694 & 2.31 \\
    Rundfunk. & 57 & 1.88 & 887 & 2.95 \\
    not\_readable& 53 & 1.75 & 3 & 0.01 \\
    Eth.Rac.Nat. & 43 & 1.42 & 884 & 2.94 \\
    Religion & 32 & 1.06 & 967 & 3.22 \\
    Spam/Scam & 32 & 1.06 & 41 & 0.14 \\
    Class & 26 & 0.86 & 804 & 2.68 \\
    Threats & 25 & 0.83 & 277 & 0.92 \\
    Phy. Con. & 15 & 0.50 & 174 & 0.58 \\
    Violence & 15 & 0.50 & 264 & 0.88 \\
    T\_Other & 14 & 0.50 & 1107 & 3.69 \\
    Suicide & 4 & 0.46 & 38 & 0.13 \\
    Ty\_Other & 2 & 0.07 & 66 & 0.22 \\
    \hline
  \end{tabular}
  \caption{\label{label_distribution_combined}
    Comparison of label frequency and percentage distribution in human-annotated and LLM-annotated datasets.
  }
\end{table}


\subsection{Age Analysis}

\begin{figure}[t]
      \includegraphics[width=2.8in]{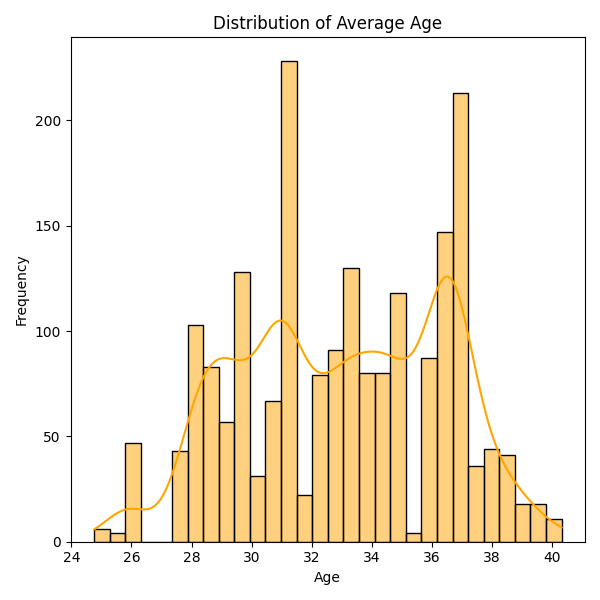}
      \caption{A figure displaying the age distribution in the 3,024 statement dataset. }
      \label{fig:age}
    \end{figure}

The average age per user is 33 years old. The overall distribution can be seen in figure \ref{fig:age}. As the age attribute was only available for Instagram and YouTube, TikTok was excluded from the age analysis. The average age for the platform Instagram is 32.56 years and for YouTube is 33.44 years. When the toxic labels are broken down into the age groups, under 30, 30-35 and over 35, see Appendix \ref{label_age_group}, Figure \ref{label_percentage_age_sorted}, the majority of content across all three age groups falls under the 'non-toxic' category. Younger individuals (0-30) show the highest percentage of non-toxic content at 84.26\%, while the 31-35 age group has the lowest at 80.26\%. This suggests a slight increase in hateful content within the middle-aged group.

To assess whether these observed differences in label distributions across age groups are statistically significant, a chi-squared test was performed. The test yielded a chi-squared statistic of 61.66 with 36 degrees of freedom and a p-value of 0.0049, indicating that there are statistically significant differences in the types of labels associated with different age groups ($p < 0.05$). These results support the hypothesis that age is associated with meaningful variation in how content is labeled, particularly regarding the presence or absence of toxic speech.

Notably, the most common type of toxic speech, “Insults,” is relatively evenly distributed across the three age groups, with percentages ranging from 8.29\% (ages 0-30) to 9.22\% (ages 31-35). In contrast, the second most frequent label, “Devaluation,” shows a significant increase in usage across age groups, being lowest among younger users (4.68\%) and highest among older users (8.08\%). For “Disinformation” and “Discrimination,” the 31-35 age group exhibits nearly double the amount of toxic content compared to users under 30, with the 35+ group falling in between.

Regarding the targets of toxic speech, “Gender and Sexual Identity” is the most frequently addressed category across all age groups. Its prevalence appears relatively consistent, with individuals aged 35+ showing a percentage of 2.91\% compared to 2.13\% in the youngest group.

Similarly, occupation-based hate is slightly more prevalent in the 31-35 age group (2.71\%) than among younger users (1.70\%). While these numbers are lower than those for insults or devaluation, they indicate that identity-based hate is persistent.

Of particular interest is that criticism of public broadcasting fees is highest in the 31-35 age group (2.28\%), decreases in the 35+ group (1.7\%), and is lowest among users aged 0-30 (0.64\%).

Religious and ethnic hate appear to be less common but still notable. Religious hate is entirely absent in the youngest age group but reaches 1.94\% among older users. Ethnic, racial, and nationality-based hate follows a similar pattern, with older users engaging in it more frequently (1.94\% for 35+ and 2.17\% for 31-35) compared to the youngest group (1.06\%). These trends suggest that younger individuals are less likely to engage in racial or religious hate speech.

Other notable findings include the slight increase in threats and spam/scams among younger users. Threats are recorded at 1.28\% for the 0-30 age group, higher than the 0.98\% observed for 31-35. Similarly, spam and scam-related content appears slightly more in the youngest demographic (1.70\%) than in the oldest (1.13\%). Mentions of violence and suicide remain rare but are slightly more present among older users.

Overall, toxicity differences by age likely reflect generational communication styles, as observed in other works \cite{Schwartz2013}. The findings indicate that younger users (0-30) exhibit the highest proportion of non-toxic content, while the middle-aged group (31-35) demonstrates the highest percentage of insults and disinformation. Older users (35+) are also more likely to spread disinformation than and the 0-30 age-group, and are more likely to engage in devaluation based on gender and sex. Although explicit violent content and threats remain relatively low across all groups, they appear in small percentages. 

\subsection{Platform Analysis}
When broken down by platform (1,000 statements per platform), see Appendix \ref{platform_analysis} Table \ref{fig:label_plat}, it shows that the highest frequency of toxic speech can be observed on Instagram (22.53\%), followed by YouTube (13.03\%), and lastly TikTok (12.43\%). Instagram, in general, has the highest amount of all types of toxic speech, mostly by a large margin, as in the case of the label “Ty: Insults,” with Instagram (10.81\%), followed by YouTube (6.82\%) and TikTok (2.30\%). 

Disinformation and Scam/Spam are also notable. 'Ty: Disinformation' appears most on Instagram (5.25\%), followed by YouTube (1.90\%) and TikTok (1.36\%), while 'Ty: Scam/Spam' follows a similar trend, being highest on Instagram (2.32\%) and lower on YouTube (0.60\%) and TikTok (0.31\%). The difference in the number of occurrences here is substantial. This suggests that Instagram has a higher presence of misleading or spam/scam content compared to the other platforms.

The only type of speech where YouTube has a stronger presence than Instagram is ‘Ty: Violence,’ but with YouTube (0.60\%), TikTok (0.52\%), and Instagram (0.41\%), they are all relatively close and not substantial in general.
For the target of toxic speech, a similar picture emerges, with Instagram dominating YouTube and TikTok. The only exception is that critique of public broadcasting fees appears most frequently on TikTok (2.50\%), followed by Instagram (1.82\%), and YouTube (1.50\%).

To assess whether these observed differences in label distributions are meaningful beyond descriptive trends, a chi-squared test of independence was conducted to test the hypothesis that content label distributions are independent of platform. The test yielded a chi-squared statistic of 269.07 with 38 degrees of freedom and a p-value less than 0.0001. This result provides strong statistical evidence to reject the null hypothesis, indicating that the differences in label distributions across platforms are not due to chance alone. Therefore platform appears to be a significant factor in determining the types and frequencies of both toxic and non-toxic content.

The findings indicate that while non-toxic content dominates all three platforms, Instagram has the highest frequency of insults, devaluation, disinformation, and discrimination. YouTube follows a similar trend but with slightly lower frequencies, while TikTok exhibits the least amount of hate speech across most categories. This may stem from platform specific design and moderation choices. Such as algorithmic amplification and the use of machine learning based moderation approaches.

\subsection{Keywords Analysis }
Appendix \ref{key_word_analyis_table}, Table \ref{label_word_frequency_human}, presents the most frequent words per age group for the labels insult and non-toxic. Emoji use is highest in the 0–30 group (15 within the top 10 words per label), followed by 31–35 (10), and 35+ (7).

For insulting language, the 0–30 group uses emojis like "\includegraphics[scale=0.03]{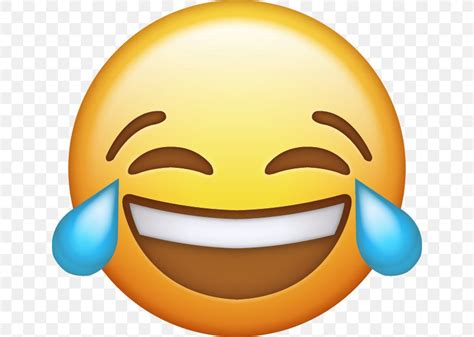}" and "\includegraphics[scale=0.04]{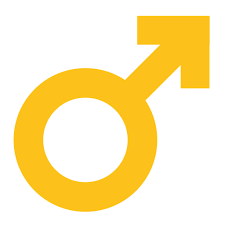}" alongside sarcastic terms like "schwachsinn" (nonsense). The 31–35 group leans into gender-based insults, frequently using "frauen" (women), "männer" (men), and "dumm" (dumb). In contrast, the 35+ group opts for broader critique, with terms like "dumm," "menschen" (humans), and "leben" (life).

In non-toxic language, the 0–30 group again favors emojis and enthusiastic terms like "geil" (awesome). The 31–35 group discusses societal themes, "menschen," "geld" (money), "video",while the 35+ group focuses on existential topics like "leben" and "gott" (God).

Overall, there is a clear difference in language used between age groups, with age group 0-30 using more emojis, while middle-aged individuals engage in more pointed social and gender-based commentary. Older individuals shift away from direct insults, using broader societal based toxicity.  



\section{Evaluation of LLM Annotated Dataset}
The dataset comprised 30,024 statements and was annotated using the fine-tuned GPT-4o-mini. Due to the lower reliability of these annotations, the analysis focuses on comparing the new annotation results with the human-annotated dataset.

The dataset includes 46 creator accounts, with an average of 652.70 comments per account. The average user age is 33.79 years, slightly higher than in the human-annotated reference set. The 30,024 collected comments were authored by 18,023 unique users. While the age distribution remains similar, the dataset contains fewer accounts. 

Examining the new label distribution in Table \ref{label_distribution_combined}, key characteristics are preserved in the LLM annotations. “Non-toxic” remains the largest class (77.02\%), though slightly decreased from 83.30\%. “Insults,” “Devaluation,” and “Disinformation” remain the next most frequent classes, with their proportions increasing in the LLM-based annotations.

The “T: Other” target class, rarely assigned in human annotations (0.50\%), became the fourth most common label in LLM annotations (3.69\%), indicating either a model error or the discovery of previously unseen hate targets in the larger dataset. 

The labels “Ty: Spam/Scam” and “Ty: Suicide,” already rare in the human-annotated dataset (1.06\% and 0.46\%), dropped sharply to 0.14\% and 0.13\%, highlighting the model’s difficulty in identifying these categories. “Violence” and “Threats” remain low, matching their human-annotated frequencies

Regarding the toxicity per age group (Appendix \ref{labels_per_age_llm}), the observations confirm previous findings. Younger users (0–30) still exhibit the highest proportion of non-toxic content, with the 31–35 group leading. The 35+ group remains in the middle. The 31–35 age group continues to have the highest scores for “Insult” and “Disinformation.” Notably, “Devaluation based on gender and sex,” previously most common in the 35+ group, is now more prevalent in the 31–35 age group.


Significant differences appear between human and LLM-based annotations at the platform level. YouTube now has the least toxicity, followed by Instagram, while TikTok ranks last by a narrow margin (see Appendix \ref{platform_analysis_LLM}). Platform-specific trends from human annotations are not reproduced, as labels are similarly distributed across platforms.


At the keyword level, results align with human annotations: the youngest age group uses the most emojis in their top 10 keywords (22), compared to 13 in the 31–35 group and 15 in the 35+ group. All results are available on GitHub.\footnote{Annonym}

The results indicate that large-scale LLM-based multi-label, multi-class annotation holds potential but also presents challenges. The LLM-based extrapolation aligns with findings based on human annotation. However, insights gathered at the platform level were not reproduced. This can indicate an error in the annotation, but it is also possible that the platform insights change within the larger dataset. Overall, the study identified four key challenges in LLM-based annotation. First, overlapping labels in multi-label tasks are a challenge. Second, scalability, accessibility, and cost are concerns, especially fine-tuning remains costly and limited compared to traditional machine learning. Third, LLM biases are unclear, influenced by both dataset choices and vague definitions of problematic speech. Finally, transparency issues, low trust due to errors, and challenges in quality assurance and evaluation present significant obstacles.






\section{Conclusions and Future Work}
This study presents the first German-language dataset that maps toxic speech across both age demographics and multiple social media platforms, developed through a collaboration between academic researchers and public broadcasting in Germany. It provides novel insights into generational and platform-specific variation in online toxicity, offering a valuable resource for both linguistic analysis and the development of more nuanced moderation systems. The research evaluated the performance of several LLMs for toxicity annotation in German, combining human and model-based approaches to assess scalability and annotation quality. Results reveal significant platform-level differences in toxic content, as well as clear age-related patterns: younger users tend to employ direct insults and expressive language, whereas older users are more likely to use disinformation and devaluative language.

While the dataset represents a step forward in multilingual and demographically-aware toxicity detection, it has several limitations, including reliance on platform-provided age data, content selection biases, misalignment of key findings between human and LLM-based annotations, and moderation discrepancies across platforms. 

Future work should incorporate context-aware toxicity modeling and explicitly address moderation dynamics to improve generalizability. Despite these challenges, this dataset represents a foundational step toward age- and platform-sensitive content moderation in German-language settings, supporting fairer, data-driven moderation strategies across diverse user populations.

\section{Limitations}

The research faces several limitations. One significant concern involves the collected age groups. Since social media platforms do not provide age data for individual users, this study estimates the age distribution of each content channel’s audience (i.e., the creator’s followers) using the demographic information form the platforms. This approach assumes that users engaging in the comment section of a channel fit the overall audience age profile, which may not always be accurate due to multi-generational viewership and the potential for users to misrepresent their age, if unintentionally or deliberately, such as to bypass platform age restrictions. As a result, the observed differences in toxic speech across age groups may not precisely reflect the actual demographics of the users producing or engaging with the content. In addition, given that the age ranges were based not on each post but on broader channel-level data, it is possible that topic differences, content style, or other confounding variables, rather than age may be driving the observed patterns.

Nevertheless, the dataset still holds substantial value for research. It remains the only available resources that incorporates any form of platform-provided demographic data, offering a level of demographic scale that is otherwise inaccessible due to privacy policies and data limitations. While the method of inference introduces uncertainty, it still allows researchers to explore broad patterns and correlations between age-associated audience characteristics and online behavior. Problems such as topic differences could be explored in future work. In general the research can serve as a foundation for further research, inform the development of more refined inference models, and guide future efforts in platform design, content moderation, and digital literacy interventions.

Another limitation is the known error within the LLM annotations. While the approach offers scalability and efficiency, is shown that errors are being made during annotation. Further, it is known that the LLMs contain biases and inconsistencies. The closed-source nature of models like GPT-4o further limits transparency, making it difficult to audit their decision-making processes or assess potential biases in classification. This raises the question: what benefit does LLM-based annotation offer if it cannot be fully relied upon? This research argues that, despite its limitations, it remains a valuable tool for data exploration and for understanding trends and insights. It is valuable as it is most likely the most cost-and time-efficient approximation of the social reality.

The classification of toxic speech also suffers from a lack of contextual understanding. Without access to full conversation threads, video content, or broader discourse context, certain comments may be misclassified.
Finally, platform-specific biases and moderation policies may affect the dataset’s representativeness. Each platform, Instagram, TikTok, and YouTube, has different moderation strategies that influence the visibility and removal of toxic content. Some platforms automatically filter out highly toxic comments before they can be collected, leading to an underestimation of certain types of online toxicity. Additionally, since TikTok does not provide age data, its role in shaping toxic speech trends remains unclear, limiting the scope of age-based analysis.

Finally, while anonymization techniques such as regex, named entity recognition, and pseudonymization were applied, these approaches are inherently imperfect. Some residual personal identifiers may remain, and the balance between data privacy and utility remains a persistent challenge in social media research. For this and other reasons, access to the dataset is restricted to scientific use under appropriate ethical and data protection protocols, ensuring responsible handling of potentially sensitive information.

\section{Ethical Considerations}
This research focuses on providing public good. Publishing these datasets for research purposes is essential to understanding fundamental societal developments. However, identifying and defining toxic speech remains a complex challenge, as it directly relates to an individual’s personal freedom of speech. The research does not claim to provide a singular definition of toxic or problematic speech, but refers to the content policies from the research collaborator funk, who is not just allowed but also obligated by law to restrict certain content within their comment sections.

To prevent misuse, it is essential to share this research and its datasets exclusively with verified scientific personnel. As a result, access to the dataset is granted only after verification. To further support reproducibility and transparency, all annotation guidelines, preprocessing scripts, and filtering criteria have been made publicly available via GitHub. Moreover, algorithmic toxic speech detection should not function independently, as relying solely on automated methods must be avoided. This research promotes a hybrid human-in-the-loop approach to online moderation, ensuring a safer digital space for all. Consequently, the ongoing collection and annotation of new datasets with human involvement remain crucial for understanding the evolving language and patterns of online hate speech.


\bibliography{custom}
\appendix
\section{Word list}
\label{wordlist}
The initial set of comments was filtered using a comprehensive word list compiled specifically to detect toxic language. This list combines terms from previously established datasets collected by the research group and published as [Annonymous]. It also encompasses a wide range of offensive and abusive vocabulary, drawing on several publicly available sources such as \texttt{Schimpfwoerter} collection on GitHub,\footnote{\url{https://gist.github.com/TheCherry/d12d53c06d134216dd404932349bdaef}} the \texttt{badwordblocker} repository,\footnote{\url{https://github.com/Uncharacteristically/badwordblocker/blob/main/bad-words.txt}} and the \texttt{swearify} dataset.\footnote{\url{https://github.com/Behiwzad/swearify/blob/master/data/words.json}} The fully used word list contains a range of vocabulary related to toxic speech and is available on GitHub.\footnote{Annonymous}

\section{Data Statement}
\label{Data_statement}
Table \ref{datastteament} displays the data statement structured as suggested by \citet{Gebru2021}. The classes “RECORDING QUALITY,” “OTHER,” and “PROVENANCE APPENDIX” were not available or applicable for the dataset.

\begin{table*}
\caption{Data statement for the dataset.}
\label{datastteament}
\begin{tabular}{p{2cm}p{12cm}}
\textbf{Characteristic}	& \textbf{Description}\\
Curation Rationale  & The dataset consists of two sets. The first contains 3,024 statements annotated by humans, while the second consists of 30,024 statements annotated by a fine-tuned GPT-4o Mini model. Both datasets were collected by funk and include comments from long- and short-format videos shared on Funk's main social media channels, as well as their associated accounts on TikTok, Instagram, and YouTube. The comments were selected based on a toxic term list. \\
Language Variety   & The messages are online, written in German. There is a visible difference in language used between age groups.\\

Speaker Demographic   & Average age of speakers are in the human annotated 3,024 sample dataset is 33 years old. And in LLM extrapolated set is 33.79 years.\\

Annotator Demographic	& Three annotators were used. They are full-time researchers with an age range between 29-45, average age 34.67. The group consisted of two males, one female. They are native German speakers. One is holding a PhD in social Anthropology and master's in Computer Science, one a master's in Psychology and Investigative Forensic Psychology, and one a master's in Information Systems. \\

Speech Situation & The dataset was collected between 01.01.2023 to 30.12.2023. It consists of written, unscripted comments under long- and short-format videos shared on funk's main social media. The intended audience were other participants of the application. \\

Text Characteristics & They are comments on funk's social media accounts. All platforms have certain moderation features in place.\\

\end{tabular}
\label{datastteament}

\end{table*}

\section{Inter-annotator Agreement Breakdown}
\label{interanno_break}
Table \ref{pairwise_percent_agreement} displays the pairwise inter-annotator agreement and Table \ref{pairwise_cohen_kappa} displays the pairwise Cohen’s $\kappa$ scores 

\begin{table}[ht]
  \centering
  \begin{tabular}{lcc}
    \hline
    \textbf{Annotator Pair} & \textbf{Agreement (\%)} \\
    \hline
    Annotator 1 vs Annotator 2 & 87.7 \\
    Annotator 1 vs Annotator 3 & 89.6 \\
    Annotator 2 vs Annotator 3 & 88.2 \\
    \hline
  \end{tabular}
  \caption{\label{pairwise_percent_agreement}
    Pairwise inter-annotator agreement percentages.
  }
\end{table}

\begin{table}[ht]
  \centering
  \begin{tabular}{lcc}
    \hline
    \textbf{Annotator Pair} & \textbf{Cohen's $\kappa$} \\
    \hline
    Annotator 1 vs Annotator 2 & 0.61 \\
    Annotator 1 vs Annotator 3 & 0.68 \\
    Annotator 2 vs Annotator 3 & 0.61 \\
    \hline
  \end{tabular}
  \caption{\label{pairwise_cohen_kappa}
    Pairwise Cohen’s $\kappa$ scores for inter-annotator agreement.
  }
\end{table}

\section{Fine-tuning LLM}
\label{fine_tune}
The tables \ref{fintuning_results}, \ref{fintuning_results_2}, and \ref{fintuning_results_3} display different settings of hyperparameters during the staged random based evaluation. 

\begin{table}
  \centering
  \begin{tabular}{p{0.05\textwidth}p{0.05\textwidth}p{0.05\textwidth}p{0.05\textwidth}p{0.05\textwidth}p{0.05\textwidth}}
    \hline
    \textbf{Batch} & \textbf{Epoch}   & \textbf{LR}& \textbf{Mac. F1} & \textbf{Acc.}   & \textbf{MCC}\\
    \hline
     3 & 5 & 0.3& 0.97 & 0.38&  0.78\\
     5 & 5 & 0.3& 0.96 & 0.33 &0.70\\
     8 & 5 & 0.3& 0.96 & 0.28 &0.68\\
    16 & 5 & 0.3&  0.95 & 0.26 & 0.63\\
    \hline
  \end{tabular}
  \caption{\label{fintuning_results}
    Results of the GPT-4o-mini fine-tuning, optimizing batch size.
  }
\end{table}

\begin{table}
  \centering
  \begin{tabular}{p{0.05\textwidth}p{0.05\textwidth}p{0.05\textwidth}p{0.05\textwidth}p{0.05\textwidth}p{0.05\textwidth}}
    \hline
    \textbf{Batch} & \textbf{Epoch}   & \textbf{LR}& \textbf{Mac. F1} & \textbf{Acc.}   & \textbf{MCC}\\
    \hline
    3 & 3 & 0.3 & 0.96 & 0.33 & 0.72\\
    3 & 5 & 0.3 & 0.97 & 0.38&  0.78\\
    3 & 8 & 0.3 & 0.97 & 0.32 & 0.73\\
    \hline
  \end{tabular}
  \caption{\label{fintuning_results_2}
    Results of the GPT-4o-mini fine-tuning, optimizing epochs.
  }
\end{table}

\begin{table}
  \centering
  \begin{tabular}{p{0.05\textwidth}p{0.05\textwidth}p{0.05\textwidth}p{0.05\textwidth}p{0.05\textwidth}p{0.05\textwidth}}
    \hline
    \textbf{Batch} & \textbf{Epoch}   & \textbf{LR}& \textbf{Mac. F1} & \textbf{Acc.}   & \textbf{MCC}\\
    \hline
    3 & 5 & 0.01 & 0.96 & 0.34&  0.71\\
    3 & 5 & 0.3 & 0.97 & 0.38&  0.78\\
    3 & 5 & 0.5 & 0.98 & 0.27 & 0.76\\
    \hline
  \end{tabular}
  \caption{\label{fintuning_results_3}
    Results of the GPT-4o-mini fine-tuning, optimizing learning rate.
  }
\end{table}

\section{Labels per Age Group}
\label{label_age_group}
The table \ref{label_percentage_age_sorted} and \ref{label_percentage_age_sorted_2} displays the percentage of labels broken down per age group. The age groups, under 30, 30-35 and over 35 where chosen due to the audience focus of the data provider funk.

\begin{table}
  \centering
  \begin{tabular}{p{0.2\textwidth}p{0.1\textwidth}p{0.1\textwidth}}
    \hline
    \textbf{Label} & \textbf{Age Group} & \textbf{Percentage} \\
    \hline
    'no\_hate' & 0-30 & 84.255319 \\
               & 35+ & 82.067851 \\
               & 31-35 & 80.260304 \\
    'Ty: Insults' & 31-35 & 9.219089 \\
                  & 35+ & 8.239095 \\
                  & 0-30 & 8.297872 \\
    'Ty: Devaluation' & 35+ & 8.077544 \\
                      & 31-35 & 7.049892 \\
                      & 0-30 & 4.680851 \\
    'Ty: Disinformation' & 31-35 & 4.555315 \\
                         & 35+ & 3.069467 \\
                         & 0-30 & 2.127660 \\
    'Ty: Discrimination' & 31-35 & 4.229935 \\
                         & 35+ & 2.584814 \\
                         & 0-30 & 1.276596 \\
    'T: Gender and Sexual Identity' & 35+ & 2.907916 \\
                                    & 31-35 & 2.711497 \\
                                    & 0-30 & 2.127660 \\
    'T: Occupation-Based' & 31-35 & 2.711497 \\
                          & 35+ & 2.100162 \\
                          & 0-30 & 1.702128 \\
    'T: Rundfunkgebühren' & 31-35 & 2.277657 \\
                          & 35+ & 1.453958 \\
                          & 0-30 & 0.638298 \\
    \hline
  \end{tabular}
  \caption{Label percentage per age group, sorted by percentage.}
  \label{label_percentage_age_sorted}
\end{table}

\begin{table}
  \centering
  \begin{tabular}{p{0.2\textwidth}p{0.1\textwidth}p{0.1\textwidth}}
    \hline
    \textbf{Label} & \textbf{Age Group} & \textbf{Percentage} \\
    \hline
    'T: Ethnic/Racial/Nationality' & 31-35 & 2.169197 \\
                                   & 35+ & 1.938611 \\
                                   & 0-30 & 1.063830 \\
    'T: Religion' & 35+ & 1.938611 \\
                  & 31-35 & 1.843818 \\
                  & 0-30 & 0.000000 \\
    'Ty: Spam/Scam' & 0-30 & 1.702128 \\
                     & 31-35 & 1.518438 \\
                     & 35+ & 1.130856 \\
    'Ty: Threats' & 0-30 & 1.276596 \\
                  & 35+ & 1.130856 \\
                  & 31-35 & 0.976139 \\
    'T: Class' & 31-35 & 1.626898 \\
               & 0-30 & 0.851064 \\
               & 35+ & 0.484653 \\
    'not\_readable' & 0-30 & 1.063830 \\
                   & 35+ & 0.646204 \\
                   & 31-35 & 0.325380 \\
    'Ty: Violence' & 35+ & 0.646204 \\
                   & 31-35 & 0.542299 \\
                   & 0-30 & 0.212766 \\
    'T: Other' & 35+ & 1.130856 \\
               & 31-35 & 0.542299 \\
               & 0-30 & 0.425532 \\
    'T: Physical Condition' & 31-35 & 0.867679 \\
                            & 0-30 & 0.638298 \\
                            & 35+ & 0.484653 \\
    'Ty: Other' & 35+ & 0.323102 \\
                & 31-35 & 0.000000 \\
                & 0-30 & 0.000000 \\
    'Ty: Suicide' & 35+ & 0.484653 \\
                  & 31-35 & 0.000000 \\
                  & 0-30 & 0.000000 \\
    \hline
  \end{tabular}
  \caption{Label percentage per age group, sorted by percentage.}
  \label{label_percentage_age_sorted_2}
\end{table}

\section{Platform Analysis}
\label{platform_analysis}
\begin{figure*}
      \includegraphics[width=6in]{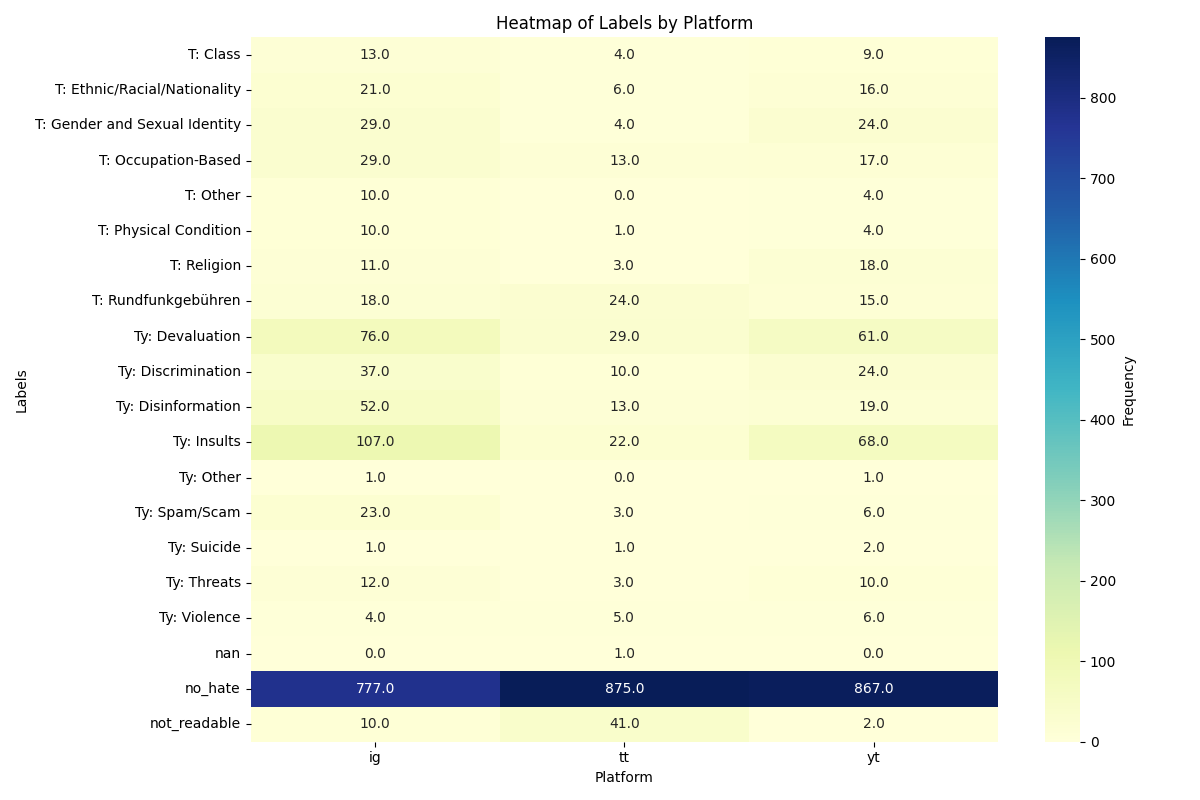}
      \caption{Heatmaps labels per platform }
      \label{fig:label_plat}
    \end{figure*}

Figure \ref{fig:label_plat} displays the labels broken down per platform.

\section{Insult and Non-Toxic Words per Age Groups.}
\label{key_word_analyis_table}
The Table \ref{label_word_frequency_human} displays the most used insult and non-toxic words for the age groups. Full keyword lists are available on GitHub.\footnote{Annonym}

\begin{table*}
  \centering
  \begin{tabular}{p{2cm} p{13cm}}
    \hline
    \textbf{Label} & \textbf{Most Used Words } \\
    \hline
    \textbf{0-30, non-tocix} & \includegraphics[scale=0.08]{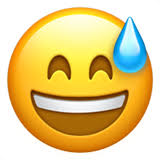}, geld (money), geil (great), video, deutschland (Germany), genau (exactly), gesellschaft (society), echt (true), \includegraphics[scale=0.08]{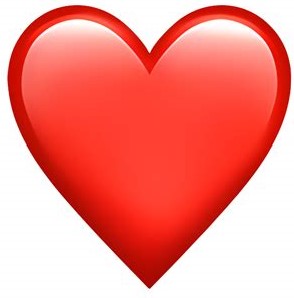}, gehört (heard) \\

    \textbf{0-30, Insults} & \includegraphics[scale=0.05]{images/emoticons/symbol_male.png}, \includegraphics[scale=0.08]{images/emoticons/laugh_crying.jpg},u200d, menschen (human), kind (kid), \includegraphics[scale=0.05]{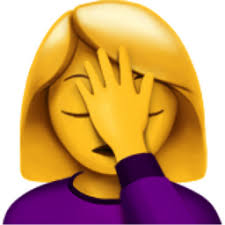}, schwachsinn (nonsense), permanent, normal, glauben (belive) \\
    \textbf{31-35, non-toxic} & menschen (human), \includegraphics[scale=0.03]{images/emoticons/laughing_crying.jpg}, video, geld (money), frauen (women), weiß (knowing/white), frage (question), halt (stop), leben (life), lustig (funny) \\
    \textbf{31-35, Insults} & frauen (women), \includegraphics[scale=0.06]{images/emoticons/face_plam_femaleDownload.jpg}, deutschland (Germany), lebt (live), nix (nothing), geld (money), bekommen (recive), \includegraphics[scale=0.03]{images/emoticons/laughing_crying.jpg} \\
    \textbf{35+, non-toxic} & menschen (human), leben (live), \includegraphics[scale=0.08]{images/emoticons/heart.jpg}, thema (topic), gott (god), frau (women), danke (thanks), interessant (interessting), weiß (knowing/white), wünsche (dreams) \\
    \textbf{35+, Ty: Insults} & \includegraphics[scale=0.1]{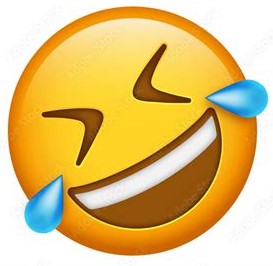}, leben (live), menschen (human), dumm (dumb), \includegraphics[scale=0.08]{images/emoticons/laugh_crying.jpg}, deutsche (german), migration, respekt (respect), kultur (culture), gesellschaft (society) \\

    \hline
  \end{tabular}
  \caption{\label{label_word_frequency_human}
    Most used insult and non-toxic words for the age group.
  }
\end{table*}

\section{Platform Analysis LLM}
\label{platform_analysis_LLM}
\begin{figure*}
      \includegraphics[width=6in]{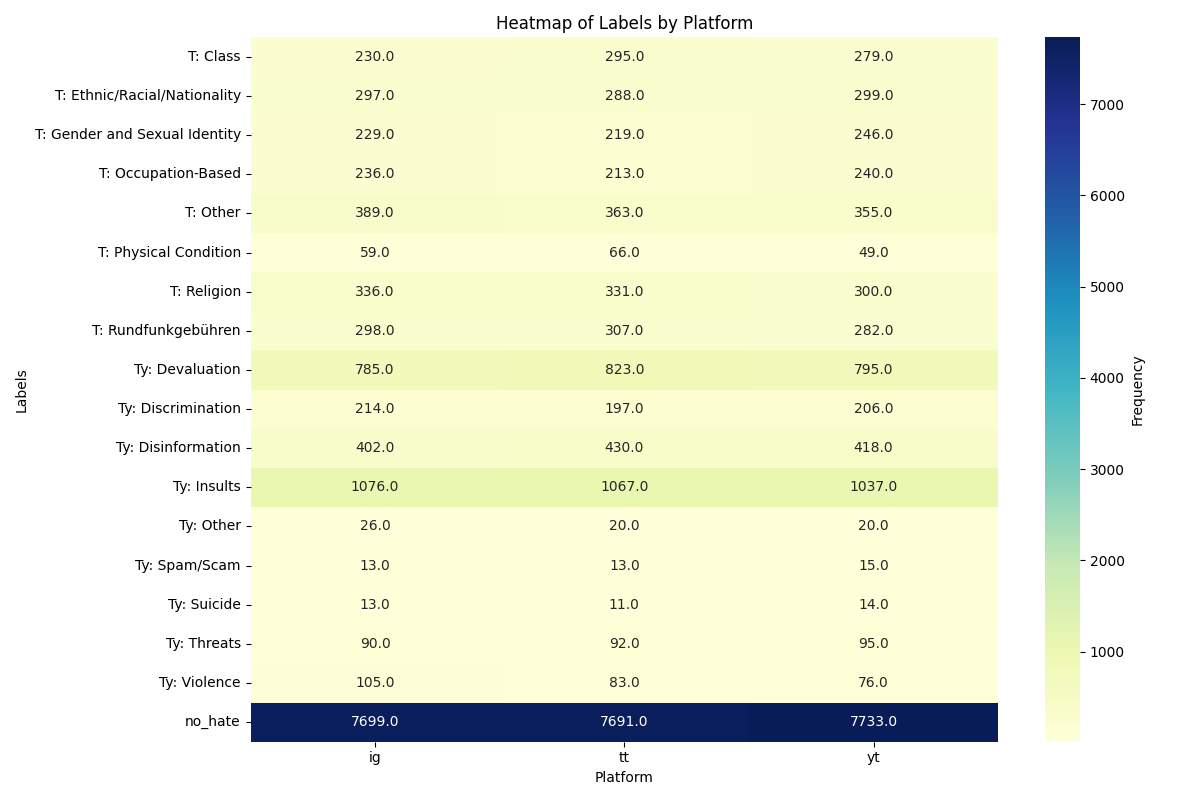}
      \caption{Heatmaps labels per platform }
      \label{fig:label_plat_LLM}
    \end{figure*}

Figure \ref{fig:label_plat_LLM} displays the labels broken down per platform.

\section{Labels per Age Group LLM Annoattion}
\label{labels_per_age_llm}
Table \ref{label_percentage_age_sorted_llm} and \ref{label_percentage_age_sorted_LLM_2} display the distribution of labels per age group.
\begin{table}
  \centering
  \begin{tabular}{p{0.2\textwidth}p{0.1\textwidth}p{0.1\textwidth}}
    \hline
    \textbf{Label} & \textbf{Age Group} & \textbf{Percentage} \\
    \hline
    'no\_hate' & 0-30 & 84.067086 \\
               & 35+ & 78.831979 \\
               & 31-35 & 73.762049 \\
    'Ty: Insults' & 31-35 & 11.012809 \\
                  & 35+ & 10.196560 \\
                  & 0-30 & 10.086187 \\
    'Ty: Devaluation' & 35+ & 7.474957 \\
                      & 31-35 & 9.441437 \\
                      & 0-30 & 4.239460 \\
    'Ty: Disinformation' & 31-35 & 5.308332 \\
                         & 35+ & 3.921754 \\
                         & 0-30 & 0.722106 \\
    'Ty: Discrimination' & 31-35 & 2.700383 \\
                         & 35+ & 1.701002 \\
                         & 0-30 & 0.652225 \\
    'T: Gender and Sexual Identity' & 35+ & 1.965602 \\
                                    & 31-35 & 2.508913 \\
                                    & 0-30 & 2.469136 \\
    'T: Occupation-Based' & 31-35 & 2.792817 \\
                          & 35+ & 2.088452 \\
                          & 0-30 & 1.048218 \\
    'T: Rundfunkgebühren' & 31-35 & 3.538888 \\
                          & 35+ & 2.362502 \\
                          & 0-30 & 2.352667 \\
    \hline
  \end{tabular}
  \caption{Label percentage per age group annotated by LLM, sorted by percentage.}
  \label{label_percentage_age_sorted_llm}
\end{table}

\begin{table}
  \centering
  \begin{tabular}{p{0.2\textwidth}p{0.1\textwidth}p{0.1\textwidth}}
    \hline
    \textbf{Label} & \textbf{Age Group} & \textbf{Percentage} \\
    \hline
    'T: Ethnic/Racial/Nationality' & 31-35 & 3.644527 \\
                                  & 35+ & 2.532603 \\
                                  & 0-30 & 1.490799 \\
    'T: Religion' & 35+ & 3.676054 \\
                  & 31-35 & 3.618117 \\
                  & 0-30 & 0.698812 \\
    'Ty: Spam/Scam' & 0-30 & 0.302819 \\
                    & 31-35 & 0.151855 \\
                    & 35+ & 0.047250 \\
    'Ty: Threats' & 0-30 & 0.815281 \\
                  & 35+ & 0.812701 \\
                  & 31-35 & 1.029975 \\
    'T: Class' & 31-35 & 3.208768 \\
               & 0-30 & 1.979967 \\
               & 35+ & 2.201852 \\
    'Ty: Violence' & 35+ & 0.897751 \\
                   & 31-35 & 0.996963 \\
                   & 0-30 & 0.419287 \\
    'T: Other' & 35+ & 3.676054 \\
               & 31-35 & 4.047273 \\
               & 0-30 & 2.445842 \\
    'T: Physical Condition' & 31-35 & 0.686650 \\
                            & 0-30 & 0.302819 \\
                            & 35+ & 0.538651 \\
    'Ty: Other' & 35+ & 0.122850 \\
                & 31-35 & 0.264096 \\
                & 0-30 & 0.302819 \\
    'Ty: Suicide' & 35+ & 0.170100 \\
                  & 31-35 & 0.072626 \\
                  & 0-30 & 0.209644 \\
    \hline
  \end{tabular}
  \caption{Label percentage per age group annotated by LLM, sorted by percentage.}
  \label{label_percentage_age_sorted_LLM_2}
\end{table}

\end{document}